\documentclass[onecolumn]{101tokens}

\usepackage{tablefootnote}
\usepackage{longtable}
\usepackage{hyperref}
\usepackage{amssymb}
\usepackage{makecell, cellspace, caption}
\usepackage{array}
\newcolumntype{L}[1]{>{\raggedright\let\newline\\\arraybackslash\hspace{0pt}}m{#1}}
\newcolumntype{C}[1]{>{\centering\let\newline\\\arraybackslash\hspace{0pt}}m{#1}}
\newcolumntype{R}[1]{>{\raggedleft\let\newline\\\arraybackslash\hspace{0pt}}m{#1}}

\usepackage[english]{babel}
\usepackage{amsmath}
\usepackage{amssymb}
\usepackage{stackengine}
\usepackage{comment}
\usepackage{pdfpages}
\usepackage[table]{colortbl}
\usepackage{wrapfig}
\setcitestyle{number,square}
\usepackage{xspace} % useful for proper spacing with variable names (see the \aya definition below)
\usepackage{lipsum}
\usepackage{blindtext}
\usepackage{array}
\usepackage{tabularx}
\usepackage{longtable}
\usepackage{xltabular}
\usepackage{placeins}
\usepackage{breqn} % For multiline equations
\usepackage{makecell} % For line breaks in table cells
\usepackage{multirow} 

\usepackage[framemethod=TikZ]{mdframed}
\usepackage{tcolorbox}
\usepackage{multicol}
\usepackage{textcomp} % for some special symbols
\usepackage{epsdice}
\usepackage{pifont}

\definecolor{ayad}{RGB}{148, 156, 229} % aya dataset #4C6EE6, opacity 60% => for text background
\definecolor{ayadsymbol}{RGB}{76, 110, 230} % aya dataset #4C6EE6, opacity 95% => for diamond symbol
\definecolor{lightblue}{RGB}{211, 227, 252} % Light blue => datacard
\definecolor{bgblue}{RGB}{247, 250, 255} % datacard background

\definecolor{ayac}{RGB}{255, 175, 71} % aya collection #FFAF47 => text background
\definecolor{lightyellow}{RGB}{250, 224, 189} % Light yellow
\definecolor{bgyellow}{RGB}{255,251,246} %light yellow for bg data card

\definecolor{ayae}{RGB}{196,178,188} % aya collection #C4B2BC => text background
\definecolor{ayaebackground}{RGB}{250, 245, 248}

\definecolor{ayaui}{RGB}{204, 232, 204} % aya dataset #CCE8CC

\newenvironment{card}[3]{
  \mdfsetup{
    frametitle={
      \tikz[baseline=(current bounding box.east),outer sep=0pt]
      \node[anchor=east,rectangle,fill=#2]{#1};
    },
    innertopmargin=7pt,
    innerbottommargin=7pt,
    innerleftmargin=7pt,
    innerrightmargin=7pt,
    linecolor=#2,
    linewidth=0.3pt,
    topline=true,
    backgroundcolor=#3,
    frametitleaboveskip=\dimexpr-\ht\strutbox\relax,
  }
  \begin{mdframed}[]\relax%
}{
  \end{mdframed}
}

\newtcolorbox{mybox}[2][]{
  colback=white, 
  colframe=lightblue,
  fonttitle=\bfseries,
  coltitle=black,  
  title=#2, 
  #1
}
\newtcolorbox{mybox2}[2][]{
  colback=white, 
  colframe=lightyellow,
  fonttitle=\bfseries,
  coltitle=black,  
  title=#2, 
  #1
}

\newtcolorbox{mybox3}[2][]{
  colback=white, 
  colframe=lightgray,
  fonttitle=\bfseries,
  coltitle=black,  
  title=#2, 
  #1
}

\DeclareSymbolFont{extraup}{U}{zavm}{m}{n}
\DeclareMathSymbol{\vardiamond}{\mathalpha}{extraup}{87}
\DeclareMathSymbol{\varspade}{\mathalpha}{extraup}{85}

\definecolor{Gray}{gray}{0.9}

\title{101 Billion Arabic Words Dataset}

\renewcommand{\fasymbol}{\textcolor{ayadsymbol}{$\vardiamond$}}% example on how to change it
\multiauthors
\author{
    name={Manel Aloui}
}
\author{
    name={Hasna Chouikhi}
}
\author{
    name={Ghaith Chaabane}
}
\author{
    name={Haithem Kchaou}
}
\author{
    name={Chehir Dhaouadi}
}

\corresponding[*] {\{manel,hasna,ghaith,haithem,chehir\}@clusterlab.ai}

\affiliations{
Clusterlab Team
}

\date{\today}
\abstract{In recent years, Large Language Models (LLMs) have revolutionized the field of natural language processing, showcasing an impressive rise predominantly in English-centric domains. These advancements have set a global benchmark, inspiring significant efforts toward developing Arabic LLMs capable of understanding and generating the Arabic language with remarkable accuracy. Despite these advancements, a critical challenge persists: the potential bias in Arabic LLMs, primarily attributed to their reliance on datasets comprising English data that has been translated into Arabic. This reliance not only compromises the authenticity of the generated content but also reflects a broader issue—the scarcity of original quality Arabic linguistic data.
This study aims to address the data scarcity in the Arab world and to encourage the development of Arabic Language Models that are true to both the linguistic and nuances of the region. We undertook a large-scale data mining project, extracting a substantial volume of text from the Common Crawl WET files, specifically targeting Arabic content. The extracted data underwent a rigorous cleaning and deduplication process, using innovative techniques to ensure the integrity and uniqueness of the dataset. The result is the 101 Billion Arabic Words Dataset, the largest Arabic dataset available to date, which can significantly contribute to the development of authentic Arabic LLMs. This study not only highlights the potential for creating linguistically and culturally accurate Arabic LLMs but also sets a precedent for future research in enhancing the authenticity of Arabic language models.

}

\begin{document}

\section{Introduction}

The emergence of Large Language Models (LLMs) has significantly advanced the field of natural language processing (NLP), transforming how machines understand and generate human-like text \citep{yao2024survey}. These models, trained on extensive datasets \citep{elnouby2021largeScale}, have primarily enhanced English language processing, leading to an increase in English-centric applications and research  \citep{valmeekam2022large}. However, the development of LLMs for other languages, such as Arabic, has not kept pace, highlighting a need for dedicated datasets \citep{Singh2024,Slim2024} and models trained from scratch to accommodate the unique linguistic features of these languages \citep{maxwell2006frontiers,joshi2019unsung}.

The dominance of English in the computational linguistics landscape has led to a technological and cultural imbalance \citep{robinson2023chatgpt,held2023material}, with languages like Arabic receiving less attention. Despite its rich literary heritage and over 400 million speakers, Arabic has been less prioritized due to a reliance on multilingual models and perceived limitations in data availability, compromising the cultural and linguistic integrity of Arabic on digital platforms. \citep{abdelali2024larabench}. The literature reveals a gap in dedicated Arabic language resources \citep{muraoka2023crosslingual}, with most studies defaulting to multilingual LLMs that inherently carry biases from dominant languages like English, thus weakening the Arabic language's unique attributes \citep{magueresse2020lowresource}.

To counter these challenges, our study introduces the \textbf{101 Billion Arabic Words Dataset}, encompassing over 101 billion words of pure Arabic content. This vast dataset challenges the existing narrative of data scarcity and aims to underpin the development of authentic Arabic LLMs. Driven by the necessity to bridge the technological divide and promote linguistic equity, our research seeks to equip the Arabic NLP domain \citep{wang2023large,abburi2023generative,li2024flexkbqa,jin2024comprehensive} with a dataset that truly reflects the language's cultural and lexical depth.

We aim to provide a comprehensive Arabic corpus that not only refutes the data scarcity argument \citep{Sengupta2023}
but also serves as a catalyst for developing Arabic-centric LLMs, thus supporting linguistic and cultural integrity in NLP technologies \citep{Lancaster2024}. The introduction of the \textbf{101 Billion Arabic Words Dataset} is a significant step toward rectifying the imbalance in language technology \citep{Jung2024}, offering a resource that enhances the development of Arabic-specific LLMs and promotes linguistic diversity and cultural authenticity.

The subsequent sections explore the methodology implemented in constructing the dataset, covering stages such as data acquisition, cleaning, and preprocessing techniques. Each step is detailed to provide insight into the processes involved in ensuring the dataset's quality and reliability. A dedicated section is allocated to discuss the limitations encountered throughout our approach, acknowledging the complexities and challenges we faced during the dataset collection while also highlighting potential avenues for future work. The paper concludes by emphasizing  the role of the dataset in advancing culturally and linguistically authentic Arabic language models. Our endeavor is driven by the aspiration to stimulate more inclusive natural language processing innovations, thereby propelling progress within the Arabic linguistic community.

\section{Related work}

The scarcity of Arabic datasets has prompted several initiatives aimed at gathering data to support progress in Arabic Natural Language Processing (NLP) research. Among such initiatives, the \textbf{KIND} dataset \citep{kind} emerges as a significant effort toward collecting nuanced dialectal data through social collaboration, a method diverging from the common practice of relying on social media extractions. Similarly, the introduction of the \textbf{ArQuAD} dataset \citep{traduct1}, a large-scale, expert-annotated corpus for Arabic machine reading comprehension, highlights the critical need for substantial, high-quality datasets to propel Arabic NLP forward.

The scarcity of datasets tailored to the Arabic language is not the sole challenge; the cultural relevance of these datasets also plays a pivotal role in the effective training of Large Language Models (LLMs). \textbf{CIDAR} \citep{cidar} addresses this gap by offering an instruction dataset that mirrors the diverse cultures across the Arab region, stepping away from the English-centric datasets that dominate the field. However, most of these emerging datasets, including \textbf{ArabicaQA} \citep{arabicaqa} and \textbf{CIDAR} \citep{cidar}, predominantly serve the purpose of fine-tuning existing models rather than facilitating the development of new models from the ground up.

This didn’t prevent some researchers from exploring monolingual models, as seen in the \textbf{AraMUS} project \citep{aramus}. Despite these efforts, the challenge of acquiring sufficient monolingual data persists, often compelling researchers to resort to English or multilingual datasets as a stopgap solution.

This reliance is exemplified in the creation of \textbf{Jais} and \textbf{Jais-chat} models \citep{jais}, which, despite their Arabic-centric foundation, incorporate a mix of Arabic and English texts in their pre-training phase. This approach underscores a prevalent belief in the field—that high-quality Arabic dataset, particularly from sources like Common Crawl\footnote{\url{https://commoncrawl.org}}, is hard to come by. Contrasting this belief, the\textbf{ RefinedWeb} \citep{refweb} dataset demonstrates that with appropriate filtering and deduplication, web data alone can yield powerful models, challenging the traditional reliance on curated corpora for training performant models with broad generalization abilities. In a parallel endeavor, we have developed an Arabic dataset following the principles highlighted by the RefinedWeb initiative. Our work not only confirms the potential of web-derived data in building models but also showcases the adaptability of these methods specifically for Arabic, indicating a meaningful advancement in creating more accessible and scalable data sourcing strategies for language model training in the Arabic linguistic domain.

\section{101 Billion Arabic Words Dataset}\label{Dataset construction and pipeline}
In this research, we explore the extensive Common Crawl archive to curate the \textbf{101 Billion Arabic Words Dataset}, analyzing 39,502 compressed WET files and extracting 0.8 petabytes (PB) of data. Common Crawl, known for its broad spectrum of content in multiple languages, often contains a mix of accurate and inaccurate information due to its comprehensive web crawling. The archive is updated monthly, providing data in raw (WARC), UTF-8 text (WET), and metadata (WAT) formats, each serving different data needs with minimal redundancy across collections. Given its 13-year history of unrestricted web crawling, the Common Crawl's text quality is varied. Our approach to crafting the 101 billion Arabic words dataset centered on meticulously selecting, filtering, and refining Arabic content from these WET files to optimize a dataset for NLP applications, thereby enhancing the resources available for Arabic language processing.

\subsection{Collection Process}\label{source}
In order to compile our dataset, we extracted a sub-corpus from the Common Crawl WET files, spanning from week 39 of 2021 to week 27 of 2022. This sub-corpus includes six distinct data splits. Each data split contains, on average, around 90,000 segments, with typically 33,292 websites, of which about 250 are in the Arabic language. 

The collection process followed the following protocol:
\begin{enumerate}
   \item Select the ID: yyyy-mm-dd representing the year, month and day for the split.
  \item Download the WET file path
  \item Open the gzip file reader.
  \item Download files.
  \item Extract only Arabic websites.
 \item Store the record in the format described below.

\end{enumerate}

\begin{figure}[!h]
 \centering
  
  \includegraphics[width=0.6\textwidth]{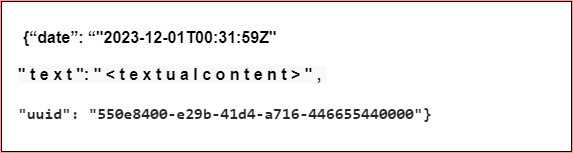}
  
  \caption{JSONL file fields.}
  \label{dedupexemple}
  \end{figure}

The resulting output from this process is formatted in JSONL, where each line represents a distinct JSON document. The corpus is distributed in this format, with each document containing the following fields in Figure \ref{dedupexemple}.

\subsection{Cleaning and preprocessing technique}\label{cleanP}
In the process of creating the \textbf{101 Billion Arabic Words Dataset}, we benefited from \textit{Yamane's formula} as a means to select a limited portion of the dataset for manual inspection. It was useful in gaining insights into the inherent noisiness of the raw text, revealing a number of issues that could potentially compromise the integrity and usability of the dataset. These issues included the presence of random HTML tags inserted between paragraphs, the use of special character Unicode, the existence of empty lines and multiple spaces, and the inclusion of adult websites and other inappropriate content. Such issues not only detract from the overall quality of the dataset but also raise concerns regarding its safety and suitability for natural language processing (NLP) tasks.

To address the challenges mentioned above and ensure the dependability and integrity of the dataset, we developed a dataset cleaning and preprocessing pipeline. This pipeline consisted of a series of carefully designed measures and procedures aimed at eliminating noise, rectifying errors, and enhancing the overall quality of the textual data.  %(see Figure \ref{pipeline}).  

%\begin{figure}[h]
 
% \centering
  %\includegraphics[width=0.85\textwidth]{pipeline_fig1.drawio.png}
  %\caption{Data Pipeline}
 % \label{pipeline}
%\end{figure}
%\begin{figure}[h]
 % \centering
 % \resizebox{0.9\columnwidth}{!}{ %
 % \includegraphics[width=1\textwidth]{fig22.png}
%}
 % \caption{An example of Text Cleaning  Pipeline.}
 % \label{fig:example}
 
%\end{figure}

\begin{itemize}
    \item \textbf{Sample Extraction} \\
    Taro Yamane’s formula (1964) provides a practical approach to estimating sample size in survey sampling, particularly when dealing with large populations. In the context of a dataset comprising 116,652,000,000 documents, applying Yamane's formula with a desired margin of error of 1\% yields a calculated sample size of approximately 10,000 documents. This calculated sample size represented a carefully determined subset of the population allowing us to draw meaningful conclusions about the entire dataset. By incorporating the population size and desired margin of error into the formula, we could efficiently design and determine the appropriate sample size needed to achieve reliable results. With a sample size of 10,000 documents, we could effectively collect and analyze data while ensuring statistical validity.

\newpage

 Yamane's formula is:
\begin{equation*}
  n= \frac{N}{1+N(e)^2}
\end{equation*}
 Given:
 \begin{itemize}
 \item Population size (N) = 116,652,000,000 documents
\item Desired margin of error (e) = 1\% (0.01)
 \end{itemize}

Using Yamane's formula, we extracted a representative subset of 10000 documents for manual inspection allowing us to identify issues such as the presence of random HTML tags, special character Unicode, empty lines and multiple spaces, adult websites, and inappropriate content.

\end{itemize}

%\newpage

\subsubsection{Filtering}\label{filtering}

\begin{figure}[h]
  \centering
  \includegraphics[width=\textwidth]{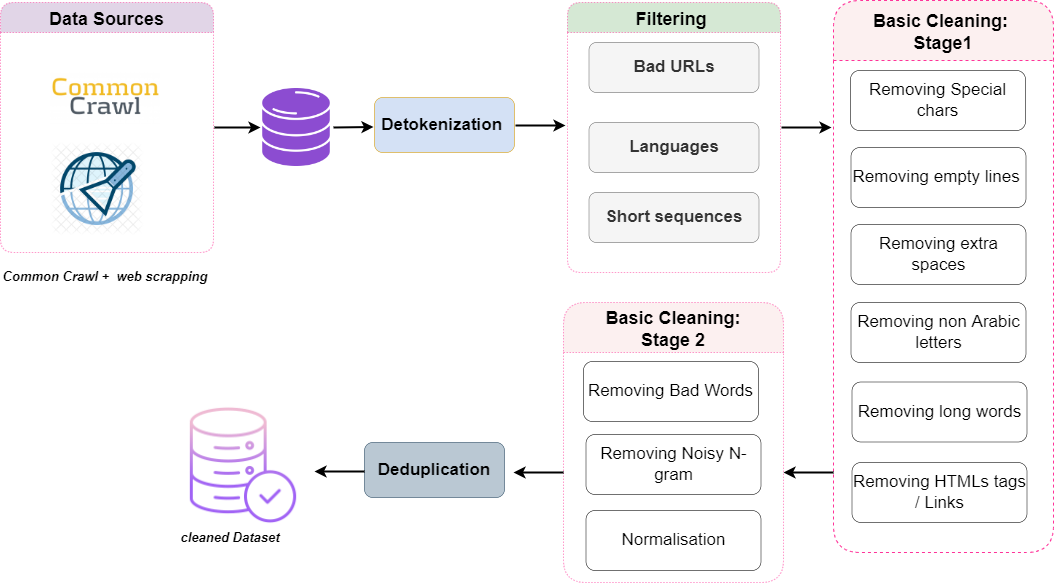}
  \caption{Dataset cleaning pipeline}
  \label{pipeline}
\end{figure}

We provide a visual in Figure \ref{pipeline} representation of the filtering Pipeline to remove noise, rectify errors, and ensure data integrity. It is crucial for NLP tasks by eliminating irrelevant content.

\begin{enumerate}[label=\alph*)]
  \item \textbf{\textit{URLs filtering:}} 
  \newline
  \newline
  Given the prevalence of low-quality machine-generated spam or adult content in Common Crawl, filtering URLs presented a significant challenge \citep{refweb}. To address this, we initiated the process by excluding URLs containing predefined inappropriate terms to filter out texts associated with such URLs. \newline
We started by using a predefined text file containing a list of inappropriate terms to scan URLs for any matches. This allowed us to identify URLs potentially containing objectionable content. Then, we used the Rust crate rustrict v 0.7.24 for further validation. This approach enabled a thorough examination of the URLs, applying additional criteria or rules beyond simply matching against the predefined list of bad words. This two-step process is ensured through the filtering of URLs.

  \item \textbf{\textit{URLs Deduplication:}} 
  \newline
  \newline
   Duplicate data can significantly detract from the quality of Large Language Models (LLMs), making URL deduplication a critical preprocessing step for maintaining the integrity and coherence of Arabic datasets. \newline To achieve this, we adopted an approach involving the use of hash representations of each URL along with the instance UUID. This technique enabled us to identify duplicated URLs accurately and precisely. By leveraging this method, we eliminated duplicate URLs effectively, reducing redundancy and improving both data storage and retrieval processes. Such measures are essential in establishing a solid foundation for subsequent NLP tasks, ensuring that models developed from the dataset are built on accurate and reliable data. 
\end{enumerate}

\subsubsection{Basic Text Cleaning}
\begin{enumerate}[label=\alph*)]
  \item \textbf{\textit{Removing Short and long Sequences}} 
  \newline
  \newline
  To refine the dataset's quality and balance, we filtered out both brief and overly long sequences. This step played a crucial role in reducing noise and filtering out unnecessary elements from the corpus, making it better suited for NLP applications like text classification and sentiment analysis. Additionally, it helped prevent data imbalance, creating a more representative and well-structured dataset for subsequent processes. 

   \item  \textbf{\textit{Cleaning of HTML Tags and Unicode}} 
    \newline
    \newline
   The process of cleaning the dataset involved removing special characters, particularly non-standard Unicode characters that might cause errors in text processing. These characters fall outside the standard UTF-8 character set and were removed to ensure accurate and consistent text analysis. For example, characters  % Using textcomp package
like \texttt{"\textbackslash u+25A1"} were omitted due to their lack of meaningful information.
\newline
The Mozilla readability library, also referred to as the reader module embedded in the most used browser, played a key role in extracting the essential content from web pages. This module extracts plain text from web pages, excluding social media citations and HTML tags. Social media citations often contain formatting specific to the platform, which can vary across different sources. Likewise, HTML tags, while essential for organizing web pages, can introduce discrepancies in the dataset. By eliminating these elements, we standardized the data in a way that is more conducive to analysis.
    \item \textbf{\textit{Empty Lines and Repeated Spaces}}
 \newline
 \newline
We removed empty lines and extra spaces to refine the dataset and improve its readability.  
The process of cleaning empty lines and repeated spaces involved identifying and eliminating consecutive line breaks and spaces between words. This was possible using regular expressions and string manipulation techniques to ensure that the resulting dataset is well-structured and consistent.

     \item \textbf{\textit{Normalizing Unicode and Dediacritization}}
 \newline
 \newline
 We refined our dataset through Unicode normalization and dediacritization, standardizing character representations and simplifying Arabic text, respectively. These processes not only enhanced dataset consistency and improved readability but also ensured uniformity in text processing across diverse encoding schemes. By normalizing Unicode, we improved integration with numerous NLP tools, promoting cross-platform compatibility. Furthermore, dediacritization optimized text for computational tasks and broadened access to Arabic language resources, enhancing engagement and collaboration in NLP applications.
 
\end{enumerate}
\subsection{Deduplication } \label{dedup}
We used MinHash to effectively deduplicate at the document level, creating hash values for parts of documents (known as "shingles")\citep{refweb} to spot nearly identical content. We also used signature-based methods at the paragraph level, generating unique signatures for sentences to help identify and eliminate repeated text. This approach was instrumental in generating compact document summaries, enabling us to quickly spot and eliminate duplicates. We enhanced this process by leveraging Rust's type system and memory safety to filter out duplicate unique identifiers (UUIDs).

Our initial dataset comprised 13.2 billion web pages from 440,000 gzip sub splits of Common Crawl, with around 110 million being Arabic websites. This large and diverse collection was then carefully refined. By focusing on URL-level uniqueness, we reduced the dataset to 99 million unique Arabic web pages. Post-deduplication, we condensed our dataset to 89.1 million web pages, amounting 0.4 terabytes (400 GB). 

\subsection{Tools}
For normalization and dediacritization, we used the Camel tools library, version 1.5.2 \citep{caml},which is a comprehensive open-source Python Toolkit for Arabic Natural Language Preprocessing created by NYU Abu Dhabi. This toolkit includes a wide range of functionalities such as preprocessing, morphological modeling, dialect identification, named entity recognition, and sentiment analysis, which were important in our text processing efforts. 

For the basic cleaning of our dataset, we leveraged the capabilities of Tnkeeh v0.0.9 \citep{tnkeeh}. This tool is tailored for the preprocessing of Arabic text and offers a suite of features to improve text quality and uniformity. Tnkeeh helps with various text cleaning tasks, such as Unicode normalization and HTML tag removal, providing a groundwork for subsequent data processing stages.
\newline
For our dataset preprocessing pipeline, we incorporated Rust to ensure memory safety and consistency checks, key for the computationally intensive NLP tasks at hand. To elevate our processing speed to unmatched levels, we used distributed computing techniques via Redis, 
The Redis pub/sub system facilitated the orchestration of task allocation and resource management, optimizing our preprocessing operations. Additionally, Redis's in-memory storage capabilities granted us immediate access to intermediate data, avoiding the slowdowns associated with disk I/O operations and substantially boosting our workflow's efficiency. This method ensured low latency, enabling our pipeline to swiftly access data needed frequently and proceed without delays through subsequent processing stages.
This approach allowed us to cut down our preprocessing time by 40 times compared to traditional techniques. 

\subsection{Environments}
To streamline the integration and processing of our dataset, we configured essential components to establish a conducive environment. This setup was crafted to promote seamless interaction between our dataset and processing tools, facilitating efficient data handling and analysis. Our method entailed the selection of compatible software tools, the enhancement of hardware resources, and the implementation of dependable data management practices.
\newline
This resulted in the creation of a cohesive system that effectively supported the complex requirements of large-scale data processing. Our architectural design ensured the dataset's accessibility and usability for a wide array of computational tasks, thereby boosting the efficiency and productivity of our research efforts. 
\begin{itemize}

 \item \textbf{Virtual Machine:}\\
We leveraged a virtual machine environment, providing scalability and flexibility to accommodate varying computational demands while maintaining cost-effectiveness.
We opted for AWS EC2 instances (see Table \ref{envConfig}).

\begin{table}[!h]
    \centering
    \resizebox{0.95\columnwidth}{!}{ %
    \begin{tabular}{cccccc}
    \hline
 \textbf{Instance Type}&\textbf{vCPU}&\textbf{Memory (RAM)}&\textbf{Storage}&\textbf{Number of instances}&\textbf{Region} \\
 \hline
 g5.4xlarge&16&64 GiB&19 TB&1&us-east-1\\
  t2.2xlarge&8&32 GiB&4 TB&10&us-east-1 \\
  \hline
    \end{tabular}
    }
    \caption{Virtual Machine Configurations.}
    \label{envConfig}
\end{table}
\item \textbf{Operating System: } \\
We configured the virtual machine with Ubuntu 22.04 LTS (64-bit) as the operating system.

\item \textbf{GPU Acceleration:} \\
We used GPU resources to accelerate data processing and analysis, enabling quicker insights and model training for large-scale datasets. To promote R\&D activities, we made use of an A10G GPU equipped with 24 VRAM, providing adequate computational power for complex tasks.

\item \textbf{Programming Language:}\\
We installed Python v3.10 and Rust v1.77.1 to support data processing and high-performance computing tasks.
We also used Anaconda distribution to manage Python packages, providing an extensive ecosystem for data manipulation and analysis.
We also used Rust for specific performance-critical operations to enhance processing efficiency in the cleaning pipeline.

\item \textbf{Cloud Storage:}\\
We configured our AWS storage volumes to ensure reliable performance during data handling and processing activities. We allocated the volumes to 16,000 Input/Output Operations Per Second (IOPS), a specification carefully chosen to boost the speed of communication between the storage units and the computational resources. This high IOPS setting was crucial for reducing latency in data-intensive operations, thus contributing to a more responsive computational environment. 

To manage potential throughput bottlenecks, we configured the system's throughput setting to 1000 Mebibytes per second (MiB/s) to ensure better distribution of bandwidth across various processes. This prevents any single task from monopolizing the system’s data transfer capabilities, thereby maintaining optimal performance and stability. This method of resource allocation contributes to enhancing overall system efficiency and supports the consistent execution of multiple simultaneous processes.

\item \textbf{Object Storage:} \\
In the context of optimizing network performance, the placement of EC2 instances and S3 buckets within the same AWS region plays critical role. Firstly, reduced latency occurs as data transfer happens within AWS’s internal network infrastructure, minimizing geographical dis- tances and reliance on the public internet. This physical proximity ensures minimal delays in data transmission. Secondly, AWS offers a network backbone that facilitates high-bandwidth connectivity, enabling faster data transfers compared to inter-region or public internet con- nections. Thirdly, AWS eliminates data transfer charges for communication between EC2 instances and S3 buckets within the same region, resulting in potential cost savings. Lastly, optimized network routing within a region, including low latency between Availability Zones, further enhances communication speed between services. Collectively, these factors contribute to the fast and efficient communication between EC2 and S3 when configured in the same AWS region, making it a recommended practice for enhancing performance and reducing costs in AWS architectures.

\end{itemize}
\section{101 Billion Arabic Words Dataset Analysis}
\begin{itemize}
    \item \textbf{Analysis of URL Filtering Efficiency:} \\
    Our findings indicated a successful URL filtering, with only 4\% of the total URLs classified as undesirable or potentially harmful. This achievement reflects the effectiveness of our filtering algorithm in accurately discerning and excluding URLs that do not align with the research criteria, thereby maintaining the dataset’s quality and relevance. 
 \item \textbf{Geographic Distribution of Filtered URLs:}\\
Investigating the domain name distribution provided insight into the geographical origins of the content within our dataset. Predominantly, the filtered URLs from Arabic domains originated in Saudi Arabia, accounting for the highest proportion within the top 10\% of our dataset. This finding not only underscores the strong online presence of Saudi Arabian domains but also indicates the importance of focusing on this region for more refined content filtering in future studies. 
\newpage
 \begin{figure}[!h]
   \centering
   %\resizebox{0.7\columnwidth}{!}{ %
  \includegraphics[width=0.98\textwidth]{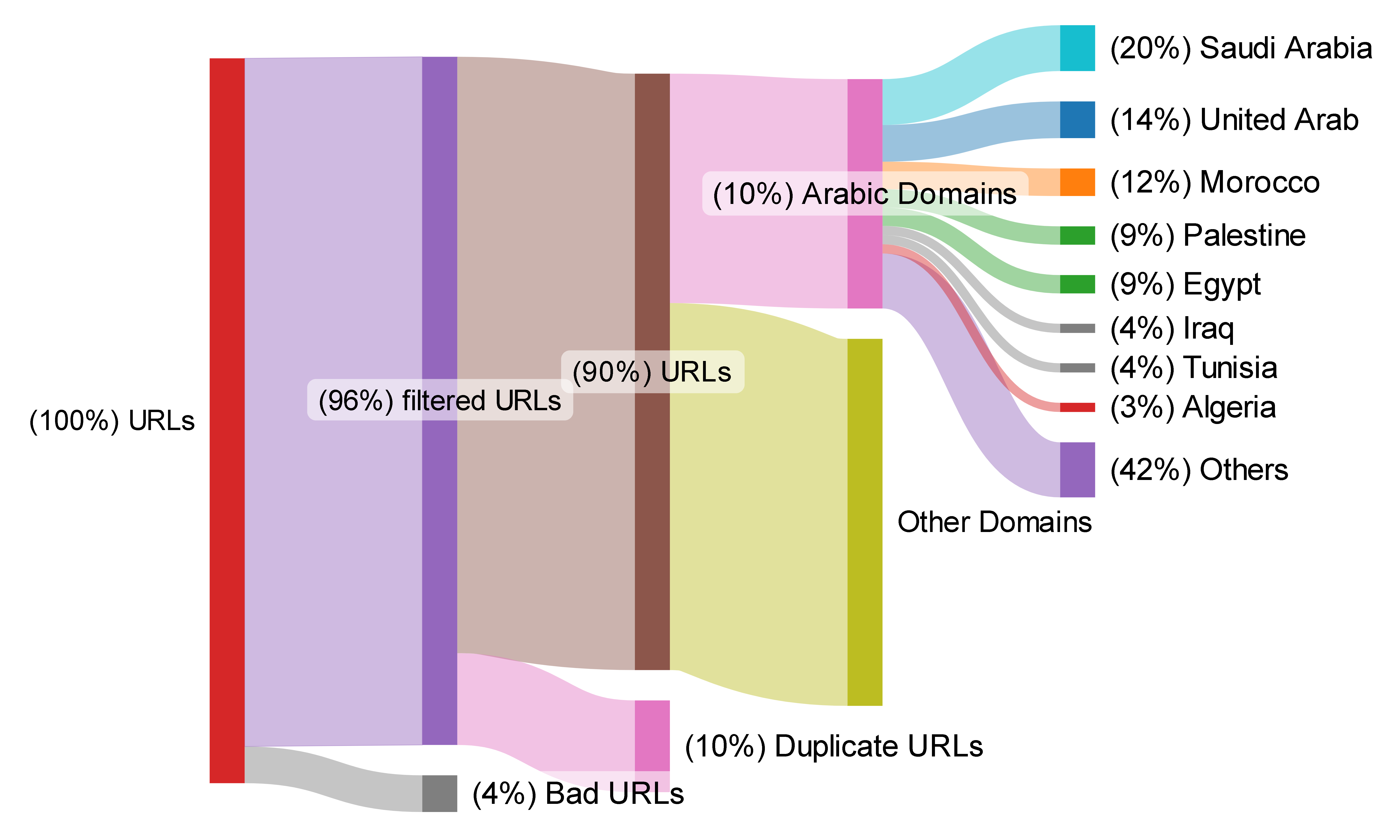}
    %}
   \caption{Distribution of filtered URLs, beginning with the initial collected dataset and narrowing down to the final dataset of 101 billion Arabic words (10\% Arabic domains)}
   \label{filtered1}
 \end{figure}
 
These findings are graphically represented in Figure \ref{filtered1}, which illustrates the distribution of filtered URLs across different geographic regions, with particular emphasis on the prevalence of Saudi Arabian domains.
The graphical representation in Figure \ref{mapDN} illustrates the geographic distribution of filtered URLs, emphasizing the significant contribution of Saudi Arabian domains. This visualization aids in comprehending the extent of regional influence on the dataset's composition. 

 \begin{figure}[!h]
   \centering
  \includegraphics[width=0.7\textwidth]{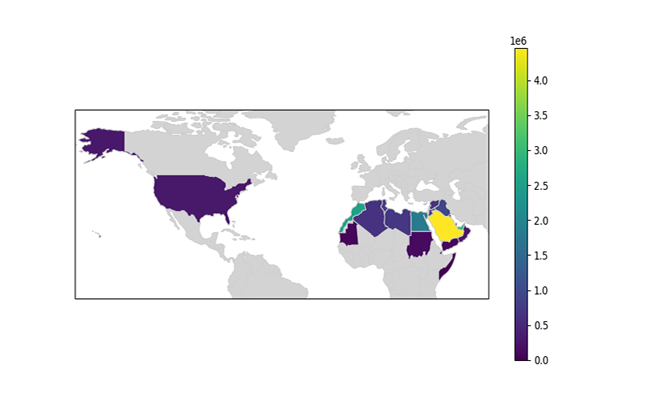}
   \caption{Geographical Distribution of Arabic Websites Categorized by Country Domain Names}
   \label{mapDN}
 \end{figure}
 \item \textbf{Content Distribution Across Domains: } \\

Figure \ref{filtered2} describes the content distribution among various domains, highlighting a noticeable scarcity of content from North African Arab countries. This imbalance points to the necessity for more sophisticated content filtering strategies that consider the regional disparities in digital content availability and representation.

  \begin{figure}[!h]
   \centering
   \resizebox{0.7\columnwidth}{!}{ %
  \includegraphics[width=0.8\textwidth]{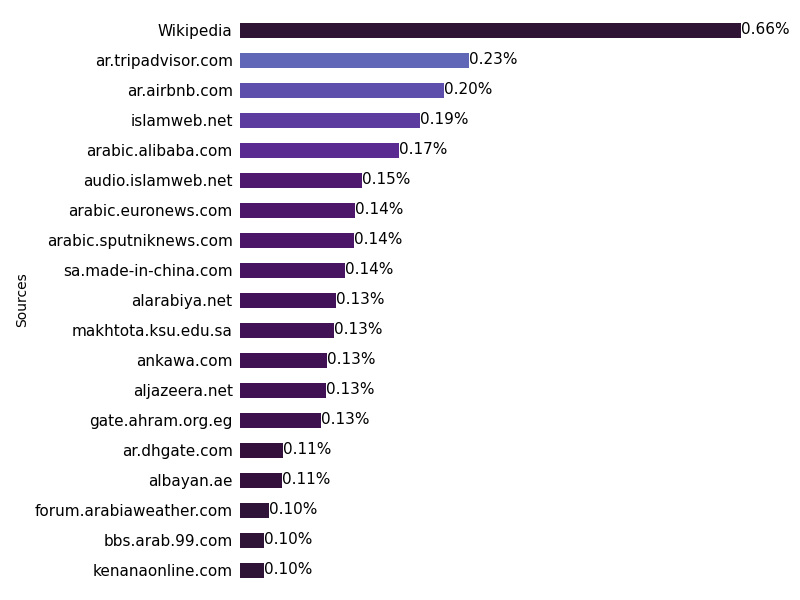}
    }
   \caption{Content distribution across various domains.}
   \label{filtered2}
 \end{figure}
 \end{itemize}

%\newpage
 
\section{Limitations}
Despite employing state-of-the-art techniques for preprocessing and assembling our dataset, a significant limitation of our research remains the lack of direct evaluation of the dataset's cleanliness and effectiveness. The most robust method to thoroughly assess these qualities involves training a model using the dataset as the primary corpus. This step would offer definitive evidence of the dataset's capacity to support the development of an effective Arabic Large Language Model that is culturally attuned. Due to computational constraints, we were unable to proceed with this crucial phase of model training. Consequently, the performance and applicability of our dataset for various NLP tasks remain untested through practical model training experiences. Acknowledging this gap is essential, as it highlights an important area for future work where, with sufficient computational resources, the dataset can be comprehensively evaluated to confirm its quality and usefulness for addressing a wide array of NLP challenges.

Ensuring the appropriateness of our dataset's content presented a significant challenge in our project. It was essential to eliminate harmful, biased, or inappropriate content for ethical utilization. Nonetheless, relying solely on standard URL filtering techniques may have limitations in capturing all sensitive subjects comprehensively.
 
% \end{itemize}
\section{Conclusion}
Our paper introduced the \textbf{101 Billion Arabic Words Dataset}, a curated corpus that is drawn from Arabic web content. 
By making this dataset accessible on HuggingFace, we extend an invitation to the international NLP community to explore its rich content, with the hope that it will drive forward research and stimulate the development of new Arabic language models. This initiative is expected to fill the gap in the availability of high-quality, large-scale Arabic language resources, thereby accelerating progress in Arabic NLP. 
Future work will focus on analyzing the dataset's diversity to systematically document the range of topics and narratives that reflect the richness of Arabic culture and language. Through this initiative, we aim to broaden the scope of available Arabic language resources, thus supporting the creation of more sophisticated and accurate language models.

\bibliography{main}

\begin{thebibliography}{29}
\providecommand{\natexlab}[1]{#1}
\providecommand{\url}[1]{\texttt{#1}}
\expandafter\ifx\csname urlstyle\endcsname\relax
  \providecommand{\doi}[1]{doi: #1}\else
  \providecommand{\doi}{doi: \begingroup \urlstyle{rm}\Url}\fi

\bibitem[Abburi et~al.(2023)]{abburi2023generative}
Harika Abburi et~al.
\newblock Generative ai text classification using ensemble llm approaches.
\newblock \emph{arXiv preprint arXiv:2309.07755}, 2023.

\bibitem[Abdallah et~al.(2024)Abdallah, Kasem, Abdalla, Mahmoud, Elkasaby, Elbendary, and Jatowt]{arabicaqa}
Abdelrahman Abdallah, Mahmoud Kasem, Mahmoud Abdalla, Mohamed Mahmoud, Mohamed Elkasaby, Yasser Elbendary, and Adam Jatowt.
\newblock Arabicaqa: A comprehensive dataset for arabic question answering.
\newblock \emph{arXiv preprint arXiv:2403.17848}, 2024.

\bibitem[Abdelali et~al.(2024)]{abdelali2024larabench}
Ahmed Abdelali et~al.
\newblock Larabench: Benchmarking arabic ai with large language models.
\newblock 2024.

\bibitem[Alghamdi et~al.(2023)Alghamdi, Duan, Jiang, Wang, Wu, Xia, Wang, Zheng, Rezagholizadeh, Huai, et~al.]{aramus}
Asaad Alghamdi, Xinyu Duan, Wei Jiang, Zhenhai Wang, Yimeng Wu, Qingrong Xia, Zhefeng Wang, Yi~Zheng, Mehdi Rezagholizadeh, Baoxing Huai, et~al.
\newblock Aramus: Pushing the limits of data and model scale for arabic natural language processing.
\newblock \emph{arXiv preprint arXiv:2306.06800}, 2023.

\bibitem[Alyafeai \& Al-Shaibani(2020)Alyafeai and Al-Shaibani]{tnkeeh}
Zaid Alyafeai and Maged Al-Shaibani.
\newblock {ARBML}: Democritizing {A}rabic natural language processing tools.
\newblock In Eunjeong~L. Park, Masato Hagiwara, Dmitrijs Milajevs, Nelson~F. Liu, Geeticka Chauhan, and Liling Tan (eds.), \emph{Proceedings of Second Workshop for NLP Open Source Software (NLP-OSS)}, pp.\  8--13, Online, November 2020. Association for Computational Linguistics.
\newblock \doi{10.18653/v1/2020.nlposs-1.2}.
\newblock URL \url{https://aclanthology.org/2020.nlposs-1.2}.

\bibitem[Alyafeai et~al.(2024)Alyafeai, Almubarak, Ashraf, Alnuhait, Alshahrani, Abdulrahman, Ahmed, Gawah, Saleh, Ghaleb, Ali, and Al-Shaibani]{cidar}
Zaid Alyafeai, Khalid Almubarak, Ahmed Ashraf, Deema Alnuhait, Saied Alshahrani, Gubran A.~Q. Abdulrahman, Gamil Ahmed, Qais Gawah, Zead Saleh, Mustafa Ghaleb, Yousef Ali, and Maged~S. Al-Shaibani.
\newblock Cidar: Culturally relevant instruction dataset for arabic, 2024.

\bibitem[El-Nouby et~al.(2021)]{elnouby2021largeScale}
Alaaeldin El-Nouby et~al.
\newblock Are large-scale datasets necessary for self-supervised pre-training?
\newblock 2021.

\bibitem[Held et~al.(2023)]{held2023material}
William Held et~al.
\newblock A material lens on coloniality in nlp.
\newblock \emph{arXiv preprint arXiv:2311.08391}, 2023.

\bibitem[Jin et~al.(2024)]{jin2024comprehensive}
Hanlei Jin et~al.
\newblock A comprehensive survey on process-oriented automatic text summarization with exploration of llm-based methods.
\newblock \emph{arXiv preprint arXiv:2403.02901}, 2024.

\bibitem[Joshi et~al.(2019)Joshi, Barnes, Santy, Khanuja, Shah, Srinivasan, Bhattamishra, Sitaram, Choudhury, and Bali]{joshi2019unsung}
Pratik Joshi, Christain Barnes, Sebastin Santy, Simran Khanuja, Sanket Shah, Anirudh Srinivasan, Satwik Bhattamishra, Sunayana Sitaram, Monojit Choudhury, and Kalika Bali.
\newblock Unsung challenges of building and deploying language technologies for low resource language communities.
\newblock pp.\  211--219, 2019.
\newblock URL \url{https://aclanthology.org/2019.icon-1.25}.

\bibitem[Jung \& van~der Plas(2024)Jung and van~der Plas]{Jung2024}
Vincent Jung and Lonneke van~der Plas.
\newblock Understanding the effects of language-specific class imbalance in multilingual fine-tuning.
\newblock 2024.

\bibitem[Lancaster(2024)]{Lancaster2024}
Thomas Lancaster.
\newblock A large language model supported synthesis of contemporary academic integrity research trends.
\newblock 2024.

\bibitem[Li et~al.(2024)]{li2024flexkbqa}
Zhenyu Li et~al.
\newblock Flexkbqa: A flexible llm-powered framework for few-shot knowledge base question answering.
\newblock 38\penalty0 (17), 2024.

\bibitem[Magueresse et~al.(2020)Magueresse, Carles, and Heetderks]{magueresse2020lowresource}
Alexandre Magueresse, Vincent Carles, and Evan Heetderks.
\newblock Low-resource languages: A review of past work and future challenges.
\newblock \emph{arXiv preprint arXiv:2006.07264}, 2020.

\bibitem[Maxwell \& Hughes(2006)Maxwell and Hughes]{maxwell2006frontiers}
Mike Maxwell and Baden Hughes.
\newblock Frontiers in linguistic annotation for lower-density languages.
\newblock 2006.

\bibitem[Muraoka et~al.(2023)]{muraoka2023crosslingual}
Masayasu Muraoka et~al.
\newblock Cross-lingual transfer of large language model by visually-derived supervision toward low-resource languages.
\newblock 2023.

\bibitem[Obeid et~al.(2020)Obeid, Zalmout, Khalifa, Taji, Oudah, Alhafni, Inoue, Eryani, Erdmann, and Habash]{caml}
Ossama Obeid, Nasser Zalmout, Salam Khalifa, Dima Taji, Mai Oudah, Bashar Alhafni, Go~Inoue, Fadhl Eryani, Alexander Erdmann, and Nizar Habash.
\newblock {CAM}e{L} tools: An open source python toolkit for {A}rabic natural language processing.
\newblock In Nicoletta Calzolari, Fr{\'e}d{\'e}ric B{\'e}chet, Philippe Blache, Khalid Choukri, Christopher Cieri, Thierry Declerck, Sara Goggi, Hitoshi Isahara, Bente Maegaard, Joseph Mariani, H{\'e}l{\`e}ne Mazo, Asuncion Moreno, Jan Odijk, and Stelios Piperidis (eds.), \emph{Proceedings of the Twelfth Language Resources and Evaluation Conference}, pp.\  7022--7032, Marseille, France, May 2020. European Language Resources Association.
\newblock ISBN 979-10-95546-34-4.
\newblock URL \url{https://aclanthology.org/2020.lrec-1.868}.

\bibitem[Obeidat et~al.(2024)Obeidat, Al-Harbi, Al-Ayyoub, and Alawneh]{traduct1}
Rasha Obeidat, Marwa Al-Harbi, Mahmoud Al-Ayyoub, and Luay Alawneh.
\newblock Arquad: An expert-annotated arabic machine reading comprehension dataset.
\newblock \emph{Cognitive Computation}, pp.\  1--20, 2024.

\bibitem[Penedo et~al.(2023)Penedo, Malartic, Hesslow, Cojocaru, Cappelli, Alobeidli, Pannier, Almazrouei, and Launay]{refweb}
Guilherme Penedo, Quentin Malartic, Daniel Hesslow, Ruxandra Cojocaru, Alessandro Cappelli, Hamza Alobeidli, Baptiste Pannier, Ebtesam Almazrouei, and Julien Launay.
\newblock The refinedweb dataset for falcon llm: outperforming curated corpora with web data, and web data only.
\newblock \emph{arXiv preprint arXiv:2306.01116}, 2023.

\bibitem[Pushkarna et~al.(2022)Pushkarna, Zaldivar, and Kjartansson]{Gebru}
Mahima Pushkarna, Andrew Zaldivar, and Oddur Kjartansson.
\newblock Data cards: Purposeful and transparent dataset documentation for responsible ai.
\newblock 2022.
\newblock \doi{10.1145/3531146.3533231}.
\newblock URL \url{https://doi.org/10.1145/3531146.3533231}.

\bibitem[Robinson et~al.(2023)]{robinson2023chatgpt}
Nathaniel~R. Robinson et~al.
\newblock Chatgpt mt: Competitive for high-(but not low-) resource languages.
\newblock \emph{arXiv preprint arXiv:2309.07423}, 2023.

\bibitem[Sengupta \& et~al.(2023)Sengupta and et~al.]{Sengupta2023}
Neha Sengupta and et~al.
\newblock Jais and jais-chat: Arabic-centric foundation and instruction-tuned open generative large language models.
\newblock 2023.

\bibitem[Sengupta et~al.(2023)Sengupta, Sahu, Jia, Katipomu, Li, Koto, Afzal, Kamboj, Pandit, Pal, et~al.]{jais}
Neha Sengupta, Sunil~Kumar Sahu, Bokang Jia, Satheesh Katipomu, Haonan Li, Fajri Koto, Osama~Mohammed Afzal, Samta Kamboj, Onkar Pandit, Rahul Pal, et~al.
\newblock Jais and jais-chat: Arabic-centric foundation and instruction-tuned open generative large language models.
\newblock \emph{arXiv preprint arXiv:2308.16149}, 2023.

\bibitem[Singh \& et~al.(2024)Singh and et~al.]{Singh2024}
Shivalika Singh and et~al.
\newblock Aya dataset: An open-access collection for multilingual instruction tuning.
\newblock 2024.

\bibitem[Slim \& Melouah(2024)Slim and Melouah]{Slim2024}
Amel Slim and Ahlem Melouah.
\newblock Low resource arabic dialects transformer neural machine translation improvement through incremental transfer of shared linguistic features.
\newblock \emph{Arabian Journal for Science and Engineering}, pp.\  1--17, 2024.

\bibitem[Valmeekam et~al.(2022)]{valmeekam2022large}
Karthik Valmeekam et~al.
\newblock Large language models still can't plan (a benchmark for llms on planning and reasoning about change).
\newblock \emph{arXiv preprint arXiv:2206.10498}, 2022.

\bibitem[Wang et~al.(2023)]{wang2023large}
Xinyi Wang et~al.
\newblock Large language models are implicitly topic models: Explaining and finding good demonstrations for in-context learning.
\newblock 2023.

\bibitem[Yamani et~al.(2024)Yamani, Alziyady, AlYami, Albelali, Albelali, Almulhim, Alsulami, Alfarraj, and Al-Zaidy]{kind}
Asma Yamani, Raghad Alziyady, Reem AlYami, Salma Albelali, Leina Albelali, Jawharah Almulhim, Amjad Alsulami, Motaz Alfarraj, and Rabeah Al-Zaidy.
\newblock The kind dataset: A social collaboration approach for nuanced dialect data collection.
\newblock In \emph{Proceedings of the 18th Conference of the European Chapter of the Association for Computational Linguistics: Student Research Workshop}, pp.\  32--43, 2024.

\bibitem[Yao et~al.(2024)]{yao2024survey}
Yifan Yao et~al.
\newblock A survey on large language model (llm) security and privacy: The good, the bad, and the ugly.
\newblock pp.\  100211, 2024.

\end{thebibliography}

\newpage

\appendix

\newpage

\section{101 Billion Arabic Words Data Card}
 Data Card for the 101 Billion Arabic Words Dataset, following the framework introduced by \citep{Gebru}, we present the dataset card for the 101 Billion Arabic Words Dataset.

\setlength{\columnseprule}{0.7pt}
\setlength{\columnsep}{20pt}

\begin{card}{\color{white} \textsf{101 Billion Arabic Words Dataset}}{ayadsymbol}{bgblue}
\small 

\label{apx:aya_dataset_datacard}

    \begin{mybox}{}
    
    {The 101 Billion Arabic Words Dataset is an Arabic language-centric dataset curated by a clusterlab. The dataset contains a total of 101 billion words for large language model training and fine-tuning.
    }

    \begin{itemize}
            \setlength\itemsep{0em}
            \item Curated by: 5 contributors from Clusterlab
            \item Language(s): Modern Standard Arabic
            \item License: Apache 2.0 \url{https://www.apache.org/licenses/LICENSE-2.0}
            \item Repository: \url{https://huggingface.co/datasets/ClusterlabAi/101_billion_arabic_words_dataset}
    \end{itemize}
    \end{mybox}

    \begin{mybox}{\textsf{Authorship}}
    \begin{multicols}{3}
        \textsf{\textbf{Publishing Organization:}}\\
        Clusterlab

        \columnbreak 

        \textsf{\textbf{Industry Type:}}\\
        Startup - Tech

        \columnbreak 

        \textsf{\textbf{Contact Details:}}\\
        {\color{red}\url{https://clusterlab.ai/}}
    \end{multicols}
    \end{mybox}

    \begin{mybox}{\textsf{Data Points}}
    
    The dataset comprises 116 million data points, providing a comprehensive and strong foundation for detailed analysis and insights.
    \end{mybox}

    \begin{mybox}{\textsf{Motivations \& Intentions}}
        The dataset was collected and gathered with the primary aim of training Arabic Language Models (LLMs) to achieve exceptional performance. It exemplifies high quality, showcasing an expansive collection of 101 billion words carefully tailored to accommodate the nuances of the Arabic language.
    \end{mybox}

        \begin{mybox}{\textsf{Provenance}}
    {
    \begin{multicols}{2}
            \textsf{\textbf{Methods Used}}\\
            We initially gathered data from specified sources, primarily Common Crawl, and extracted Arabic content from WET files using Rust. Then, we applied our preprocessing pipeline, which included text cleaning and deduplication (Figure \ref{pipeline}).
            
            \columnbreak %

            \textsf{\textbf{Methodology Details}}\\
            \textsf{Source:} mainly Common Crawl \\
            \textsf{Dates of Collection:} September 2021 - July 2022
            
            \columnbreak 
        
    \end{multicols}
    }
    \end{mybox}

       \begin{mybox}{\textsf{Dataset Version and Maintenance}}

    \begin{multicols}{3}
            \textsf{\textbf{Maintenance Status}}\\
            Actively Maintained
            
            \columnbreak 

            \textsf{\textbf{Version Details}}\\
            Current version: 1.0

            \columnbreak 

            \textsf{\textbf{Maintenance Plan}}\\
            Updates will be periodically made available.
        
    \end{multicols}

    \end{mybox}
\end{card}

\end{document}